\documentclass[11pt]{article}
\usepackage[T1]{fontenc}
\usepackage{amsthm, amsmath, amssymb, amsfonts, url, booktabs, tikz, setspace, fancyhdr, bm, mathrsfs}
\usepackage{geometry}
\usepackage{hyperref, enumerate}
\usepackage[shortlabels]{enumitem}
\usepackage[babel]{microtype}
\usepackage[english]{babel}
\usepackage[capitalise]{cleveref}
\usepackage{comment}
\usepackage{bbm}
\usepackage{algorithm}
\usepackage{algpseudocode}
\usepackage{booktabs}
\usepackage{csquotes}
\usepackage{mathabx}
\usepackage{graphicx}
\usepackage{float}
\usepackage{subcaption}

\counterwithin{figure}{section}

\theoremstyle{definition}

\newtheorem*{defn-non}{Definition}

\newlist{Case}{enumerate}{3}
\setlist[Case, 1]{%
    label           =   {\bfseries Case \arabic*.},
    labelindent=1em ,labelwidth=1cm, labelsep*=1em, leftmargin =!
}
\setlist[Case, 2]{%
    label           =   {\bfseries Subcase \arabic{Casei}.\arabic*.},
    labelindent=-1em ,labelwidth=1cm, labelsep*=1em, leftmargin =!
}
\setlist[Case, 3]{%
    label           =   {\bfseries Subsubcase \arabic{Casei}.\arabic{Caseii}.\arabic*.},
    labelindent=-1em ,labelwidth=1cm, labelsep*=1em, leftmargin =!
}

\usepackage{todonotes} 
\usepackage{cite}

\title{Active Learning for Multi-class Image Classification}

\author{
Vo Thien Nhan \thanks{
Institute of Engineering, Ho Chi Minh City University of Technology (HUTECH), Vietnam \\  Email: thiennhan.math@gmail.com}}

\begin{document}
\maketitle
\begin{abstract}
A principle bottleneck in image classification is the large number of training examples needed to train a classifier. Us- ing active learning, we can reduce the number of training examples to teach a CNN classifier by strategically selecting examples. Assigning values to image examples using differ- ent uncertainty metrics allows the model to identify and se- lect high-value examples in a smaller training set size. We demonstrate results for digit recognition and fruit classifi- cation on the MNIST and Fruits360 data sets. We formally compare results for four different uncertainty metrics. Fi- nally, we observe active learning is also effective on simpler (binary) classification tasks, but marked improvement from random sampling is more evident on more difficult tasks. We show active learning is a viable algorithm for image classi- fication problems.
\end{abstract}
\section{Introduction}
Active learning is a machine learning framework in which the learning algorithm can interactively query a user (teacher or oracle) to label new data points with truth labels. The motivation for active learning are scenarios where there is a large pool of unlabelled data. An example is training an image classification model to distinguish cats and dogs. There are millions of images of each, but a smaller size of informative samples are needed to train a good model. Similarly, this applies to problems like classifying YouTube video content, where there are many videos and training is expensive. Passive learning, the standard framework where a large quantity of labelled data is used to train, requires significant effort in labelling the large data set. Through active learning, we can selectively leverage systems like crowd-sourcing, to ask human experts to selectively label some items in the data set. The algorithm iteratively selects examples based on some value, or uncertainty metric, and has an oracle label these examples.

In several cases, active learning outperforms than ran- dom sampling. The below graph shows a motivating toy example of active learning’s improvement over random selection. The entire set of data points (union of the sets of red triangles and green circles) is not linearly separable. 

The active learning framework selects data based on what data points are the most informative, which are the ones that the model is most uncertain about. This leads to developing various methods to determine the uncertainty of examples. Active learning is motivated by the fact that not all labelled examples are equally important. With uniform random sampling we indiscriminately choose examples. In contrast, active learning selects examples typically near the class boundary, and can train a more representative classifier with less training samples
\begin{figure}[H]
    \centering
    \includegraphics[width= 0.75\linewidth]{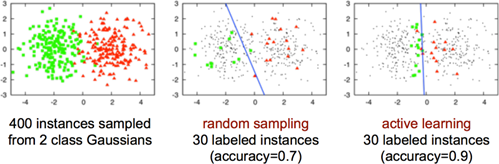}
    \caption{Illustrative Example. Active learning enables better than random selection in classification problems.}
    \label{fig:1}
\end{figure}
The primary contributions of this paper are (1) an ac- tive learning algorithm that can perform multi-class classification problems, (2) a formal comparison of four uncertainty measures, and (3) an investigation of active learning on simple versus complex classification tasks. 

\section{Related Work}
Many image classification tasks in computer vision require a large amount of training data. Active learning is selectively querying human experts for labelled data, and this framework could drastically reduce the amount of data needed to learn an accurate model. Training machine learn- ing models can be expensive due to the number of training examples needed. Active learning reduces the number of examples by choosing the most informative to train on. Specifically, we have seen that in image classification several techniques have been shown to perform effectively.

Tong et al. \cite{b7} proposes use of a support vector ma- chine active learning algorithm for conducting relevance
 
feedback for image retrieval. Their algorithm selects the most informative images to query a user using SVM (Support Vector Machine) margins for unlabeled examples, but only investigates the binary classification setting. Tong et al. \cite{b8} uses active learning to minimize a version space of tasks in binary classification. Li et al. \cite{b11} applies an adaptive active learning approach to object recognition that employs a combination of uncertainty measure and information density measure. Holub et al. \cite{b10} proposes an entropy-based active learning metric for object recognition. Hoi et al. \cite{b9} introduced batch active learning with support vector machines for the purpose of image retrieval, where the primary classification task is binary. We utilize the entropy measure introduced by Holub et al. \cite{b10} in the batch active learning setting, but formally expand upon the uncertainty measures explored in multi-class active learning for image classification.

Joshi et al. \cite{b6} extends active learning to the multi- class setting, following the idea of uncertainty sampling. Discrete entropies are measured from the class probabilities generated from the probabilistic outputs of the SVM. The two uncertainty measures compared are entropy and best-versus-second-best (B-SB) margin, where entropy outperformed B-SB. In our work, we investigate best-versus-worst (B-W) margin, and least-confidence uncertainty. Although several different methods have been shown to reduce the number of training samples needed, there hasn’t been a formal comparison done on these methods. We aim to compare these and explain their differences.

\section{Active Learning}
Active learning is a semi-supervised learning method, between unsupervised being using 0\% of the learning exam- ples and fully supervised being using 100\% of the examples. By iteratively increasing the size of our labelled training set, we can achieve greater performance, near fully-supervised performance, with a fraction of the cost to train using all of the data. In pool-based sampling (the subsection of active learning we are investigating), training examples are cho- sen from a large pool or unlabelled data. Selected training examples from this pool are labelled by the oracle.

\section{Uncertainty Measures}

The decision for selecting the most informative data points is dependent on the uncertainty measure used in se- lection. In pool-based sampling, the active learning algo- rithm selects examples to add to the growing training set that are the most informative. The most informative exam- ples are the ones that the classifier is the least certain about. The intuition behind selecting the most uncertain examples is that by obtaining the label for those particular examples, the examples with which the model has the least certainty are the most difficult examples, the most likely the ones near the class boundaries. The learning algorithm will likely gain the most information about the class boundaries by observ- ing the difficult examples. In this paper, we compared four uncertainty measures.

\subsection{Largest Margin Uncertainty}

\[
\phi_{LM}(x) = P_\theta(y_1^*|x) - P_\theta(y_{\min}^*|x)
\]

The largest margin uncertainty is a best-versus-worst uncertainty comparison. The Largest Margin Uncertainty (LMU) is the classification probability of the most likely class minus the classification probability of the least likely class. The intuition behind this metric is that if the probability of the most likely class is significantly greater than the probability of the least likely class, then the classifier is more certain about the example’s class membership. Likewise, if the probability of the most likely class is not much greater than the probability of the least likely class, then the classifier is less certain about the example’s class membership. The active learning algorithm will select the example with the minimum LMU value.

\subsection{Smallest Margin Uncertainty}
\[
\phi_{SM}(x) = P_\theta(y_1^*|x) - P_\theta(y_2^*|x)
\]

The smallest margin uncertainty is a best-versus-second-best uncertainty comparison. The Smallest Margin Uncertainty (SMU) is the classification probability of the most likely class minus the classification probability of the second most likely class. The intuition behind this metric is that if the probability of the most likely class is significantly greater than the probability of the second most likely class, then the classifier is more certain about the example’s class membership. Likewise, if the probability of the most likely class is not much greater than the probability of the second most likely class, then the classifier is less certain about the example’s class membership. The active learning algorithm will select the example with the minimum SMU value.

\subsection{Least Confidence Uncertainty}
\[
\phi_{LC}(x) = 1 - P_\theta(y_1^*|x)
\]

Least Confidence Uncertainty (LCU) is selecting the example for which the classifier is least certain about the selected class. LCU selection only looks at the most likely class, and selects the example that has the lowest probability assigned to that class.
\subsection{Entropy Reduction}
\[
\phi_{ENT}(x) = - \sum_{y} P_\theta(y|x) \log P_\theta(y|x)
\]

Entropy is the measure of the uncertainty of a random variable. In this experiment, we use Shannon Entropy.

Shannon entropy has several basic properties, including
(1) uniform distributions have maximum uncertainty, (2) uncertainty is additive for independent events, and (3) adding an outcome with zero probability has no effect, and (4) events with a certain outcome have zero effect. Considering class predictions as outcomes, we can measure Shannon entropy of the predicted class probabilities.

Higher values of entropy imply more uncertainty in the probability distribution. In each active learning step of the algorithm, for every unlabelled example in the training set, we compute the entropy over the predicted class probabilities, and select the example with the highest entropy. The example with the highest entropy is the example for which the classifier is least certain about its class membership.

\section{Algorithm} 
The algorithm iteratively selects the most informative ex- amples based on some value metric and sends those unla- belled examples to a labelling oracle, who returns the true labels for those queried examples back to the algorithm.

\begin{algorithm}
\caption{Pool-based Active Learning}
\begin{algorithmic}[1]
\State $\epsilon \gets$ training error bound
\State Divide data into unlabelled pool $P$ and test set $S$
\State Split training pool into batches
\State Randomly select $k$ examples from training pool to initialize training set $T$
\While{Training Error $> \epsilon$}
    \State Train the model using $T$
    \State Use the trained model with the test set to get performance measures
    \For{each $e \in P$}
        \State Compute uncertainty for $e$
    \EndFor
    \State Select $k$ most-informative samples based on uncertainty metric
    \State Move these $k$ examples to training set
    \State Remove these $k$ examples from pool $P$
\EndWhile
\end{algorithmic}
\end{algorithm}
\section{Experiments}
\subsection{MNIST Dataset}
\begin{figure}[H]
    \centering
    \includegraphics[width=0.5\linewidth]{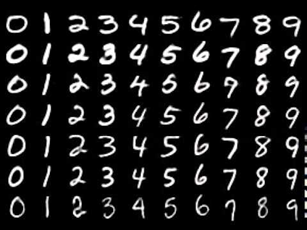}
    \caption{MNIST Dataset of Handwritten Numbers}
    \label{fig:2}
\end{figure}

The MNIST (Modified National Institute of Standards and Technology) Database is a large database of hand writ- ten digits, zero through nine. This database is widely used for training and testing machine learning models for image classification. It was created by taking and remixing sam- ples from the NIST’s original data sets. This was done since the NIST’s training data set was taken from United States Census Bureau employees and the test set was taken from high school students from the United States. The black and white images from NIST were normalized to fit into a 28x28 pixel bounding box. The MNIST database contains 60,000 training images and 10,000 testing images, which makes it a quite sizable data set. Due to these characteristics, the MNIST data set is commonly used to test out machine learn- ing models due to its small input size (28x28 gray scale pix- els) along with its large data set size.

\begin{figure}[H]
    \centering
    \includegraphics[width=0.65 \linewidth]{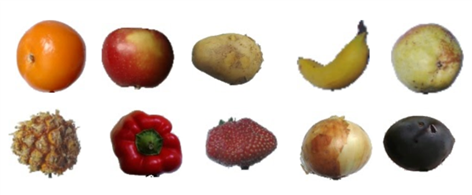}
    \caption{Fruits360 Dataset Classes with Examples}
    \label{fig:3}
\end{figure}
\subsection{Fruits360 Dataset}
The Fruits360 data set is a data set of fruits and vegeta- bles by Mendeley Data. Fruits and vegetables were planted in the shaft of a low speed motor (3 rpm) and a short movie of 20 seconds was recorded. A Logitech C920 camera was used for filming the fruits. The images show the fruits ro- tated in different angles. We made modifications to the data set in order to reduce the number of classes from 120 to
10. We selected 10 classes: Apple, Tomato, Potato, Banana, Pear, Pineapple, Pepper, Strawberry, Onion, and Plum. We combined specific strains of fruit like Crimson Snow Ap- ple and Red Delicious Apple all into the Apple category, for each fruit category.

\section{Results} 
For our experiments, the metric that we used was accu- racy, or the percentage of samples classified correctly. Our procedure is that we go through the data set and while doing so we run each batch through our model. By doing so we see which batches are the most uncertain according to the metric that we are using. From these batches, we only kept the most uncertain or informative samples and trained on these. We kept repeating this process for a specified number of epochs. We chose to resample from the training set every time because as the model trains

\subsection{MNIST Results}
The neural network model we used for MNIST was a convolutional neural network (CNN) with two covolutional layers, a pooling layer, and two fully connected layers. We also regularize with batch normalization. The MNIST database has 60,000 training images and 10,000 training images over 10 classes. The batch size for training was 128 samples with 25 batches being used for training each epoch. This means that 3,200 samples are being used every epoch for training, which is 5.3\% of the total training set selected per epoch.
\begin{figure}[H]
    \centering
    \includegraphics[width=0.5\linewidth]{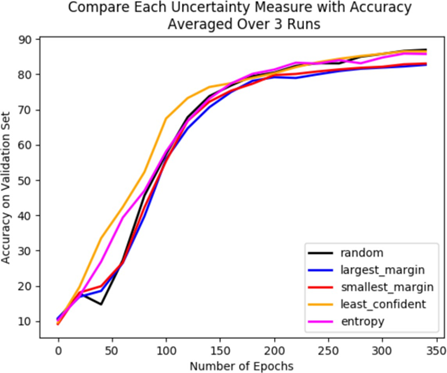}
    \caption{MNIST Dataset Active Training Set Accuracy Averaged Over Three Runs for Credibility}
    \label{fig:4}
\end{figure}

\begin{figure}[H]
    \centering
    \includegraphics[width=0.5\linewidth]{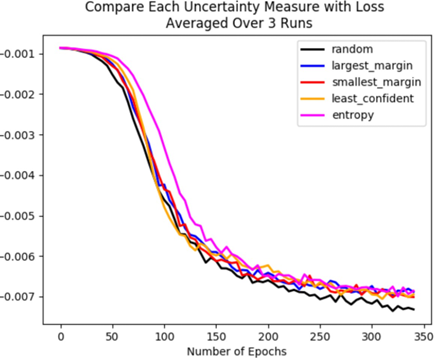}
    \caption{MNIST Dataset Active Training Set Loss Averaged Over Three Runs for Credibility}
    \label{fig:5}
\end{figure}

Analyzing our results for the multi-class classification task (with numbers zero through nine), we can see in Figure
4 (Compare Each Uncertainty Measure with Accuracy Averaged Over 3 Runs) that the Least Confident active learning method does the best, followed by the Entropy active learning method. We can see especially from epochs 25-150, that least confident outperformed all of the other active learning metrics, followed by the entropy active learning metric. Additionally, we can see that the random sampling baseline is outperformed by everyone up to 50 epochs, but actually manages to do better than the largest margin and smallest margin active learning methods 
 
It is expected that the active learning measures would have the largest training performance gain in earlier epochs since active learning aims to choose the most informative samples to train. This is seen in the graph since also from epoch 0 to 150, the curve for all active learning metrics is steepest, with least confident leveling off first. In concert with this, when we look at Figure 5, we can see that from epochs 0 to 150 that the loss per batch is steepest and then levels off there after, indicating that for our model (on the MNIST data set) most of the learning occurs from epochs 0 to 150.

Interestingly, we can see that the random sampling base- line outperforms largest margin and smallest martin active learning method after approximately epoch 70. This may be the case since these two active learning metrics are ill infor- mative of the task. For instance, largest margin only cares about the margin between the most and the least probable prediction. This does not seem to help with classification of multiple classes since it is only the most probable class that produces results and knowing the difference between the most and least likely class is non-informative when there’s eight other classes in between. Smallest margin also did not perform well. This may be because although you are look- ing at the two most likely classes, if one class gets confused with other as being highly probable (such as 7’s and 1’s or 6’s and 9’s), then this metric does not work.

On the MNIST data set, where we are trying to classify ten different classes, least confident does best followed by entropy. One can surmise that least confident does well because it trains on the data that the model is the most uncertain about. We can see this because for least confident the samples that are chosen are the ones where the selected class was the most uncertain with respect to his peers. Thus when we train on these samples we optimize for training on hard to distinguish pieces of data on the edges of classes of data. Similarly, entropy reduction is highest when we are very uncertain across all classes what class the data is, and is lowest when one class definitively is certain the data is part of its class and the rest of the classes are certain that the data is not part of their class. These two methods do the best because they choose data where we are most uncertain about the classification choice we chose.
\begin{figure}[h]
    \centering
    \includegraphics[width=0.5\linewidth]{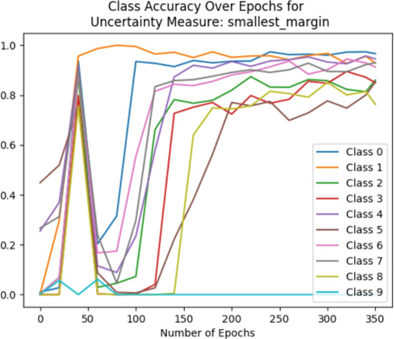}
    \caption{MNIST Dataset Smallest Margin Active Learning; Accuracy Among Classes}
    \label{fig:6}
\end{figure}

\begin{figure}[H]
    \centering
    \includegraphics[width=0.5\linewidth]{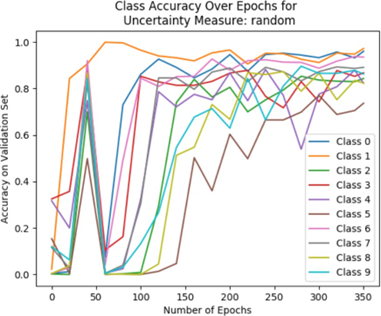}
    \caption{MNIST Dataset Random Active Learning Accuracy Among Classes}
    \label{fig:7}
\end{figure}
Also interestingly, if we peruse the graphs that depict the class accuracy over epochs (entropy, largest margin, and smallest margin included in Appendix for brevity) for vari- ous uncertainty measures, we see that for every uncertainty measure every class gets better at being classified. This is great because this means that active learning is helping to improve the classification of each class. However, this is oddly not the case for the smallest margin active learning method, as can be seen with class 9, where it has zero per- cent accuracy after about epoch 50. We believe this phe- nomenon occurs because with smallest margin we are com- paring the difference between the best and second best pre- dictions for data and choosing the data with the smallest difference between them to train on. What could have hap- pened, is that all 9’s could’ve been misclassified as another class (perhaps the similar looking 6’s) but the second most probable class predicted was deemed unlikely, so those sam- ples were never chosen for training.

\begin{figure}[H]
    \centering
    \includegraphics[width=0.5\linewidth]{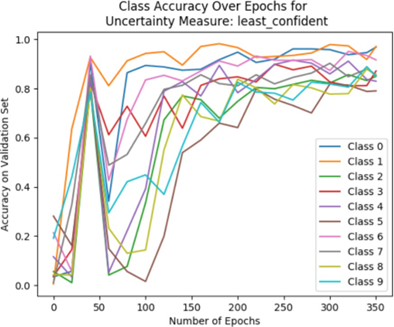}
    \caption{MNIST Dataset Least Confident Active Learning Accuracy Among Classes}
    \label{fig:8}
\end{figure}

\begin{figure}[H]
    \centering
    \includegraphics[width=0.5\linewidth]{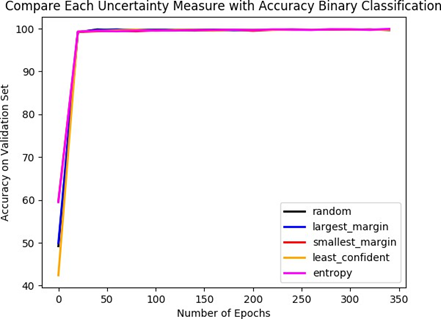}
    \caption{MNIST Dataset Binary Classification Accuracy Classes: 0 and 1}
    \label{fig:9}
\end{figure}

\begin{figure}[H]
    \centering
    \includegraphics[width=0.5\linewidth]{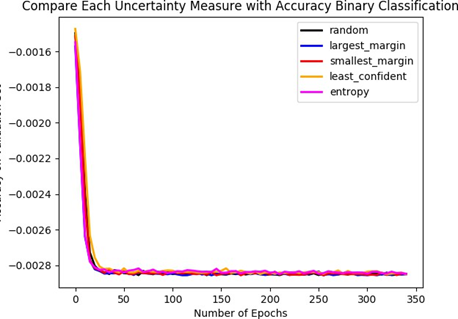}
    \caption{MNIST Dataset Binary Classification Accuracy Classes: 0 and 1}
    \label{fig:10}
\end{figure}

\begin{figure}[H]
    \centering
    \includegraphics[width=0.5\linewidth]{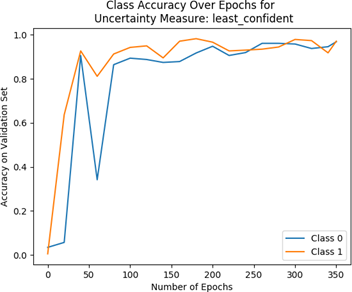}
    \caption{MNIST Dataset Binary Classification
Least Confident Active Learning; Accuracy Among Classes
}
    \label{fig:11}
\end{figure}
For completeness of our results we chose to also run ac- tive learning on a binary classification task, where we had the algorithm train to discern between zero’s and one’s. As we can see, Since this task was much easier (two classes in- stead of ten), we get that in every circumstance the model trains much faster, such as can be seen by the MNIST Dataset Binary Classification Accuracy for Classes 0 and 1 graph (smallest margin and largest margin included in Ap- pendix for brevity), where we achieve near one hundred per- cent accuracy within fifty epochs (more like twenty-five ac- tually). We can also see from the graphs depicting individ- ual uncertainty metrics that both classes within 100 epochs start to converge after 100 epochs, with least confident hav- ing the shallowest classification dip for class zero at around 50 epochs, that seems to be characteristic of training.

\begin{figure}[H]
    \centering
    \includegraphics[width=0.5\linewidth]{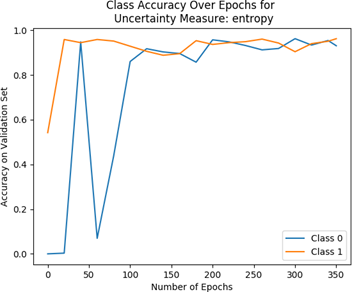}
    \caption{. MNIST Dataset Binary Classification Entropy Active Learning; Accuracy Among Classes
}
    \label{fig:12}
\end{figure}

\begin{figure}[H]
    \centering
    \includegraphics[width=0.5\linewidth]{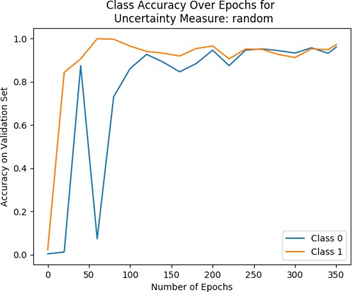}
    \caption{MNIST Dataset Binary Classification Random Active Learning; Accuracy Among Classes
}
    \label{fig:13}
\end{figure}

 \subsection{Fruits360 Results}
The active learning model used was a convolutional neu- ral network (CNN) with three convolutional layers, a pool- ing layer, and three fully-connected layers. We also regu- larize with batch normalization. The dataset contained over 30,000 fruit images distributed over 10 classes. The batch size for training was 64, and then number of batches to- tal was 372. We performed 20 epochs with 1280 training samples selected per epoch, which was selecting 4.2\% of training data per epoch.

Active learning for all uncertainty metrics outper- formed random selection. Active learning with smallest- margin (best-versus-second-best) and least confidence out- performed the other uncertainty measures entropy and largest-margin. Smallest-margin (best-versus-second-best) and least confidence performed similarly, with either one surpassing the other at times. Metrics entropy and largest- margin performed similarly. While there is a distinction be- tween the two clusters of cluster (1) smallest-margin and least confidence training quicker than cluster (2) of entropy and largest-margin, all metrics performed similarly well. This drastic improvement over random selection indicates that any sort of intelligent heuristic inserted into sample se- lection offers significant speed-up benefit in training time.

 On the test set, active learning with smallest-margin (best-versus-second-best) and least confidence also outper- formed the other uncertainty measures and outperformed random by a large margin.

An extension of our work on multi-class classification
 
was comparing the performance of active learning on multi-class classification to binary classification tasks. For the two classes, we reduced the Fruits data set to 2 classes: apples and bananas. We found that all four active learning uncertainty sampling measures outperformed random sampling. With binary classification, we found that all models could learn the binary classification task with high accuracy within one epoch, so all four active learning uncertainty sampling measures performed similarly. This shows that active learning is effective on simpler tasks, but really deviates from random sampling on more difficult tasks.

\begin{figure}[H]
    \centering
    \includegraphics[width=0.5\linewidth]{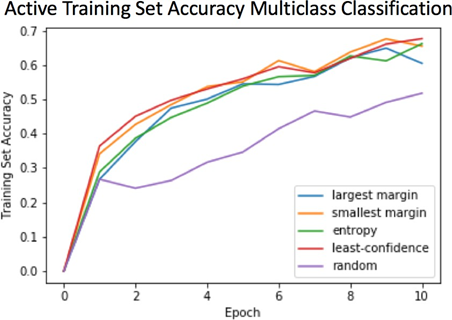}
    \caption{Results of Multi-class Active Learning on Training Set. Active learning with smallest-margin (best-versus-second- best) and least confidence outperformed the other uncertainty mea- sures and outperformed random by a large margin.
}
    \label{fig:14}
\end{figure}
\begin{figure}[H]
    \centering
    \includegraphics[width=0.5\linewidth]{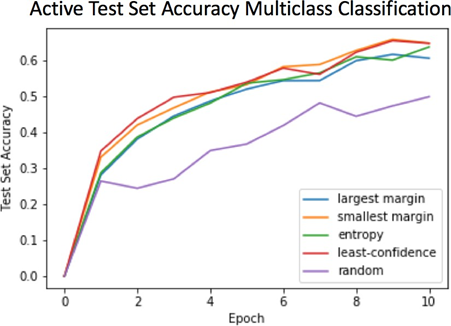}
    \caption{Results of Multi-class Active Learning on Test Set. Ac- tive learning with smallest-margin (best-versus-second-best) and least confidence outperformed the other metrics.
}
    \label{fig:15}
\end{figure}
\section{Discussion}
The first result of our work is in showing that active learning is a viable framework for multi-class image clas- sification problems. The second result of our work is that of the four uncertainty metric compared in our analysis, small- est margin (best-versus-second-best) and least-confidence gave the best performance in terms of reduction in train- ing set size for learning on the Fruits360 data set. For the MNIST data set we found that least confident followed by entropy performed the best. The third result of our work is that active learning is highly effective on simpler tasks, but marked improvement over random sampling is evident in more difficult tasks. Training with both random sampling and active learning uncertainty sampling on binary classifi- cation learns an accurate model in about one epoch (for the Fruits 360 data set), showing that active learning uncertainty sampling does not offer significant improvement over ran- dom sampling. Similarly, with MNIST, we saw that for the binary classification task, all uncertainty metrics have simi- lar convergence times. In the multi-class classification set- ting, the improvement over random sampling is significant. We see that with harder classification problems (multi-class over binary and Fruits 360 data set over MNIST) that active learning tends makes a larger difference since here it is more important what samples we train on.
 
\section{Conclusion}
Active learning allows us to address a principle bottle- neck in image classification - the large number of training examples needed for training a classifier. Our work shows that active learning reduces the number of training examples needed to teach a CNN classifier by strategically selecting informative examples. By assigning value to image exam- ples using different uncertainty metrics, the model is able to reach high test set accuracy in a significantly smaller num- ber of training examples. We demonstrate results for the two different classification tasks digit recognition and fruit clas- sification on data sets from MNIST and Fruits360. We for- mally compare results for four different uncertainty metrics, and conclude that smallest margin and entropy reduction are the two most effective uncertainty measures for Fruits 360 and that least confident and entropy are the most effective for MNIST. Finally, we observe that active learning is most effective on simpler classification tasks. By showing that active learning gives improved results on two vastly differ- ent data sets, our work gives definitive support that active learning is a viable algorithm for general image classifica- tion problems.

\begin{figure}[H]
    \centering
    \includegraphics[width=0.5\linewidth]{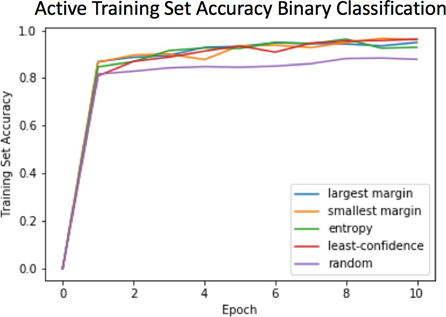}
    \caption{Results of Multi-class Active Learning on Training Set. Active learning with smallest-margin (best-versus-second- best) and least confidence outperformed the other uncertainty mea- sures and outperformed random by a large margin.
}
    \label{fig:16}
\end{figure}

\section*{Appendix and Additional Figures}

Here we are going to put additional graphs that we gen- erated for this report, but were not pertinent to the main dis- cussion.

\begin{figure}[H]
    \centering
    \includegraphics[width=0.5\linewidth]{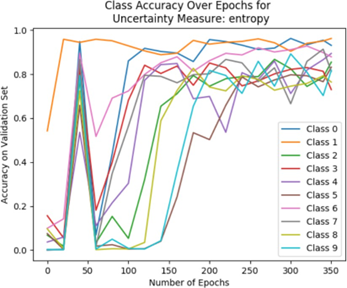}
    \caption{MNIST Dataset Entropy Active Learning Accuracy Among Classes
}
    \label{fig:17}
\end{figure}

\begin{figure}[H]
    \centering
    \includegraphics[width=0.5\linewidth]{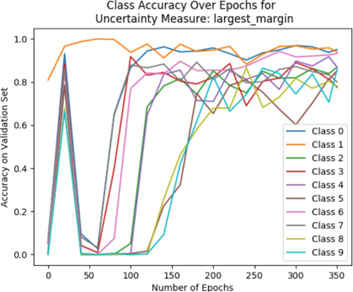}
    \caption{MNIST Dataset Largest Margin Active Learning Accuracy Among Classes
}
    \label{fig:18}
\end{figure}

\begin{figure}[H]
    \centering
    \includegraphics[width=0.5\linewidth]{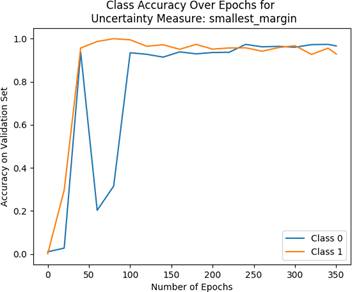}
    \caption{MNIST Dataset Binary Classification
Smallest Margin Active Learning; Accuracy Among Classes}
    \label{fig:19}
\end{figure}

\begin{figure}[H]
    \centering
    \includegraphics[width=0.5\linewidth]{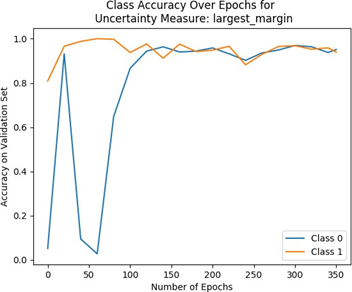}
    \caption{MNIST Dataset Binary Classification
Largest Margin Active Learning; Accuracy Among Classes
}
    \label{fig:19}
\end{figure}

\end{document}